\documentclass[journal,onecolumn]{IEEEtran}

\usepackage{cite}
\usepackage{amsmath,amssymb,amsfonts}
\usepackage{algorithmic}
\usepackage{algorithm}
\usepackage{graphicx}
\usepackage{textcomp}
\usepackage{xcolor}
\usepackage{multirow}
\usepackage{float}
\usepackage{hyperref}
\usepackage{comment}
\usepackage{booktabs}
\usepackage{siunitx}
\usepackage[nameinlink]{cleveref}
\usepackage{caption}
\begin{document}

% Paper title
\title{Continual Learning for non-stationary regression via Memory-Efficient Replay }

% Author names and affiliations
\author{Pablo García-Santaclara,
        Bruno Fernández-Castro, Rebeca~P.~Díaz-Redondo and Martín Alonso-Gamarra
        
\thanks{P. García-Santaclara and R. P. Díaz-Redondo are with atlanTTic - ICLAB - Universidade de Vigo, Escola de Enxeñaría de Telecomunicación, Campus Universitario, Vigo 36310, Spain (e-mail: pgarcia@alumnos.uvigo.es; rebeca@det.uvigo.es).}

\thanks{B. Fernández-Castro is with Centro Tecnolóxico de Telecomunicacións de Galicia (GRADIANT), Carretera do Vilar, 56-58, Vigo 36214, Spain (e-mail: bfernandez@gradiant.org).}

\thanks{M. Alonso-Gamarra is with the Facultade de Informática, Universidade de Santiago de Compostela, Campus Vida, 15782 Santiago de Compostela, Spain, (e-mail: martin.alonso.gamarra@rai.usc.es).}

}

% Make the title area
\maketitle

\begin{abstract}
Data streams are rarely static in dynamic environments like Industry 4.0. Instead, they constantly change, making traditional offline models outdated unless they can quickly adjust to the new data. This need can be adequately addressed by continual learning (CL), which allows systems to gradually acquire knowledge without incurring the prohibitive costs of retraining them from scratch. Most research on continual learning focuses on classification problems, while very few studies address regression tasks. We propose the first prototype-based generative replay framework designed for online task-free continual regression. Our approach defines an adaptive output-space discretization model, enabling prototype-based generative replay for continual regression without storing raw data. Evidence obtained from several benchmark datasets shows that our framework reduces forgetting and provides more stable performance than other state-of-the-art solutions.
\end{abstract}


% \endif

% Make the title appear
\IEEEpeerreviewmaketitle

% Sections
\section{Introduction}
\label{sec:intro}
Continual learning addresses the challenge of adapting machine learning models to new data as it changes over time, without losing the knowledge of what was previously learned. The phenomenon of losing already acquired knowledge, known as catastrophic forgetting, is a key problem in the development of adaptive systems. While the majority of research on continual learning has focused on classification problems, the challenge of maintaining performance on previously learned regression functions while adapting to new data distributions remains under-explored. In fact, this is a relevant gap, since regression tasks are important for operational and industrial systems where predictions directly support decision-making processes. Consequently, continual learning regression is very well-suited for Industry 4.0 contexts, as manufacturing environments generate data that is constantly changing. Also, these scenarios are often characterized by limited computational resources and memory capacities. Thus, incremental training becomes more suitable than traditional approaches that require large datasets and extensive computing power and storage resources. 

In this paper, we propose a novel framework to address this topic. Our work builds upon and extends TRIL3 (Tabular-data Rehearsal Incremental LifeLong Learning)~\cite{garcia2025overcoming}, a framework originally designed for continual classification of tabular data. While TRIL3 effectively addresses catastrophic forgetting in classification through a pseudo-rehearsal strategy utilizing prototype generation, applying this directly to regression presents challenges due to the continuous nature of target variables, as they lack discrete class labels to guide prototype creation and neighborhood definition. Furthermore, identical input regions may correspond to different target values over time, further complicating the notion of prototype identity. Finally, we applied a two-stage experimental procedure to validate our proposed framework. First, we evaluate the framework on a variety of heterogeneous tabular datasets against a generic replay strategy and an offline Random Forest Regressor model. Second, we perform a direct comparison with a framework created for the same purpose, CLeaR~\cite{he2021clear}. 

Our results show that our approach achieves competitive performance compared to offline baselines and outperforms experience replay in mitigating forgetting in the majority of evaluated datasets. In comparison with the similar alternative (CLeaR) our strategy demonstrates a clear advantage, having lower forgetting ratios in most of the scenarios and without performance degradation during update phases in most of the instances.

The paper structure is as follows. \Cref{sec:related} reviews the fundamental concepts of continual learning and examines the state of the art in rehearsal strategies and regression tasks.  \Cref{sec:background} provides the necessary background on the TRIL3 framework and the prototype-based algorithms that are the basis for our approach. \Cref{sec:methodology} details our methodology, introducing the integration of Mixture Density Networks for continual regression. \Cref{sec:results} presents the experimental validation, divided into two stages: a comparison against standard baselines and a direct evaluation against the CLeaR framework. Finally, \Cref{sec:conclusions} summarizes our main conclusions and outlines future research directions.

\section{Related Work}
\label{sec:related}

In this section, we first review some approaches developed to address catastrophic forgetting in neural networks, and then focus on solutions that use pseudorehearsal techniques and regression-centered studies. Research about continual learning in regression contexts is still very limited compared to CL applied to classification problems, where most studies focus on image problems rather than tabular data or time series.~\Cref{subsec:cl} describes a general vision of continual learning state-of-the-art. In~\Cref{subsec:clr}, we examine some of the most relevant rehearsal studies, as our approach is based on a pseudo-rehearsal technique. Finally,~\Cref{subsec:soa_regression} presents an overview of recent advances in continual learning for regression tasks.
\subsection{Continual Learning}
\label{subsec:cl}
Literature presents different taxonomies for continual learning strategies. Both Maltoni and Lomonaco~\cite{maltoni2019continuous} and Farquhar and Gal~\cite{farquhar2019unifying} propose a classification consisting of 3 families of techniques:

\textbf{Architectural techniques} involve modifying the model's structure during the learning process by adding components for each new task. Progressive Neural Networks~\cite{Rusu2016} freeze previously learned columns and add new ones for fresh data, to ensure zero interference. Other approaches, such as Dynamically Expandable Networks (DEN)~\cite{yoon2017lifelong}, optimize this process by selectively expanding the network capacity only when necessary. While this isolates new knowledge effectively, it leads to a significant increase in model size and computational complexity as tasks accumulate. More recently, Kang et al~\cite{pmlr-v162-kang22b} presented a framework that gradually finds task-specific winning subnetworks within an overparameterized model using binary masks. This allows for selective parameter reuse while freezing previously chosen weights to avoid forgetting.

\textbf{Regularization techniques} introduce penalty terms to the loss function to prevent drastic changes in weights that are important for previous tasks. While beneficial because it requires no extra memory storage, it creates a trade-off between stability (long-term memory) and plasticity (learning new tasks), often resulting in lower retention compared to other methods when tasks vary significantly. EWC (\textit{Elastic Weight consolidation})~\cite{kirkpatrick2017overcoming} was one of the studies introducing this technique. After training on a task, EWC determines the importance of each parameter using the Fisher Information Matrix, when learning a new task, the model decreases the learning on the weights important for the previous tasks. Gomez-Villa et al.~\cite{Gomez-Villa_2022_CVPR} propose a technique consisting of projecting updates onto subspaces that maintain previously learned behaviors in order to limit the functional changes of self-supervised representations across tasks. Kumar et al.~\cite{kumar2024maintainingplasticitycontinuallearning} present a conceptual framework that prevents excessive rigidity in long task sequences by periodically relaxing accumulated constraints. Despite these developments, regularization-based methods may still have trouble in task-free scenarios and highly non-stationary environments, where adaptation to new tasks may be limited by the preservation of prior knowledge.

\textbf{Replay strategies}, also known as rehearsal, are based on storing a subset of previous data to reuse during training. Using stored data has some challenges, such as not being suitable for privacy-critical scenarios or contexts with a lot of memory constraints, like Edge or IoT devices. Replay is generally categorized into two main approaches: experience replay and generative replay (also known as pseudo-rehearsal). The former involves storing a subset of raw training samples from previous tasks in a memory buffer to be use with new data during training. The latter avoids storing raw data and, instead, it trains a generative model to synthesize past knowledge, mitigating privacy concerns and heavy storage costs.

\subsection{Continual Learning Rehearsal}
\label{subsec:clr}

Pseudo-rehearsal, also known as generative replay, addresses catastrophic forgetting by using generative models to synthesize data from previous tasks instead of storing original samples.

The foundational work from Shin et al.~\cite{shin2017continual} introduced a dual-model architecture consisting of a generator and a solver trained cooperatively. The generator produces synthetic samples every time it learns a new task, the samples are then paired with corresponding responses from the solver, which are mixed with real data from the current task. Van de Ven et al.~\cite{van2020brain} extended this with brain-inspired replay, addressing scalability by replaying a fixed number of samples per iteration regardless of previous tasks.
Recent works like \textit{DDGR}~\cite{gao2023ddgr} introduce a bidirectional relationship where the classifier guides sample generation, demonstrating advantages in class-incremental settings. \textit{SDDGR}~\cite{kim2024sddgr} utilizes pre-trained text-to-image diffusion networks with iterative refinement. For industrial applications, \textit{DSG}~\cite{he2024continual} applies diffusion-based replay to streaming sensor data. \textit{Latent Generative Replay}~\cite{stoychev2023latent} trains generators to produce low-dimensional latent features, enabling deployment on resource-constrained devices. 

Variational autoencoders (VAEs) offer an alternative for pseudo-rehearsal. Ye and Bors~\cite{ye2022task} propose a dual-memory framework where Short-Term Memory and Teacher modules preserve long-term information through continual generative knowledge distillation. \textit{BooVAE}~\cite{egorov2021boovae} learns aggregated posterior approximations using boosting-like component addition, improving the choice of prior distribution over the latent space in a VAE model.
\textit{MalCL}~\cite{park2025malcl} employs GANs with feature matching loss for malware classification, where image-optimized CL techniques fail. Recent work addresses GAN instability through teacher-student architectures~\cite{ali2024cfts}.

\subsection{Continual Learning in regression tasks}
\label{subsec:soa_regression}

The first attempts to use continual learning for regression tasks appeared recently. The most relevant contribution is \textit{CLeaR}~\cite{he2021clear}, which introduces a repetition-based adaptive framework designed for continual prediction. The work demonstrates that techniques originally developed for classification can be reused for numerical prediction streams.\textit{CLeaR} is capable of detecting when there is a change in the distribution of the input data or a change in the mapping between inputs and outputs. When this occurs, it stores the data in a \textit{buffer} of new data. If no substantial change in the input data is identified, it is stored in a known data buffer. When the new knowledge memory is full, the system is updated by training with both the new and known stored data. Although this solution has promising results, memory limitations can be a problem, especially when working with large volumes of data or in long-term continual learning scenarios. Furthermore, storing real data can raise privacy and security issues, especially in sensitive applications.  Building on this foundation, the same author proposed an \textit{Adaptive Explainable Continual Learning Framework for Regression Problems}~\cite{he2108adaptive}, the approach enables models to incrementally extend their knowledge through adaptive learning mechanisms, with particular emphasis on non-stationarity detection and explainability. The framework is evaluated using real-world power generation and consumption forecasting tasks. The work highlights challenges in continual regression and proposes a modular, explainable framework designed to generalize across dynamic application domains.

A parallel line of research provides a mathematical characterization of forgetting in sequential regression. Ding et al.~\cite{ding2024understanding} analyze how forgetting emerges when regression models are trained with stochastic gradient descent, showing that task ordering, parameter drift, and step-size selection jointly determine forgetting magnitude. Related theoretical works explore these ideas in further depth. Evron et al.~\cite{evron2022catastrophic} derive worst-case bounds on forgetting when tasks exhibit strong dissimilarity in linear regression settings. Zhao et al.~\cite{zhao2024statistical} study continual regression under ridge-based regularization and analyze conditions under which regularization enables positive transfer. Zhu et al.~\cite{zhu2025global} extend convergence theory to sequential regression under heterogeneous input distributions. Although these studies provide insightful explanations about why and how forgetting happens, they do not offer specific solutions to reduce forgetting in real-world regression systems.

Pham et al.~\cite{pham2024certified} proposed a framework capable of training neural regression models incrementally while maintaining provable bounds on prediction error after each update. This work reflects growing industrial requirements: CL systems used in manufacturing or industrial control must ensure robustness and accountability. Samuel et al.~\cite{samuel2025continual}, compare continual learning methods in multistage engineering design tasks and conclude that traditional classification-oriented strategies (EWC, GEM) do not transfer effectively, with errors accumulating across tasks.

Continual learning regression has received much less attention than continual learning in the field of classification. The CLeaR framework provides support for continual regression by identifying the occurrence of distributional shifts and adjusting the model accordingly. However, these frameworks rely on storing actual data and updating the model through parameter-based adjustments. As such, they have limited scalability, can create privacy issues, and directly link memory growth to the length of data streams. Other work in this area emphasizes a complementary aspect to continual regression, such as providing examples of research on explainable adaptive regression systems or presenting theoretical analyses of catastrophic forgetting in sequential regression research (i.e., examples of why and how forgetting occurs, but without providing practical methods for mitigating it). In contrast to these data-driven, parameter-regularized, and theoretical approaches based on isolation, we present a solution to catastrophic forgetting using a memory-efficient method based on prototypes. Our approach replaces buffer-based replay with a generative model, where past knowledge is summarized through a compact representation. This design eliminates the need to store actual historical data entirely, allowing memory usage to stay limited. Our proposal also has other advantages of using synthetic data generation, such as avoiding privacy and security issues related to storing raw historical data.

\section{Background}
\label{sec:background}
Our approach extends TRIL3 (Tabular-data Rehearsal Incremental LifeLong Learning)~\cite{garcia2025overcoming}, a framework for continual classification of tabular data that uses a prototype generation model for the generation of synthetic data. TRIL3 effectively addresses catastrophic forgetting in classification scenarios, but regression tasks present different challenges due to the continuous nature of target variables, which are addressed in the next section. In this section, we provide an overview of TRIL3's architecture and detail XuILVQ~\cite{gonzalez2022xuilvq}, the prototype-based algorithm that forms the foundation of our regression framework and requires significant adaptations for this extension for regression tasks.

\subsection{TRIL3: Architecture and Workflow}
\label{sec:tril3_overview}

Our previous proposal, the TRIL3 framework~\cite{garcia2025overcoming}, employs a pseudo-rehearsal strategy, combining XuILVQ~\cite{gonzalez2022xuilvq}, a model whose purpose is generating synthetic data to preserve past knowledge, and Deep Neural Decision Forests (DNDF)~\cite{kontschieder2015deep}, a model that combines deep neural networks representational power with decision trees hierarchical structure that performs the classification task in the framework. 

TRIL3's primary innovation is its capacity to dynamically balance newly received data with artificial representations of previously learned patterns. When new data is received, XuILVQ updates a set of prototypes online and captures the key features of the data distribution. After that, synthetic samples are created using these prototypes and combined with fresh data to avoid catastrophic forgetting. A simplified diagram is depicted in the~\Cref{fig:simple}. The TRIL3 workflow consists of six main steps:

\begin{figure}[ht]
  \centering
    \includegraphics[width=0.8\textwidth]{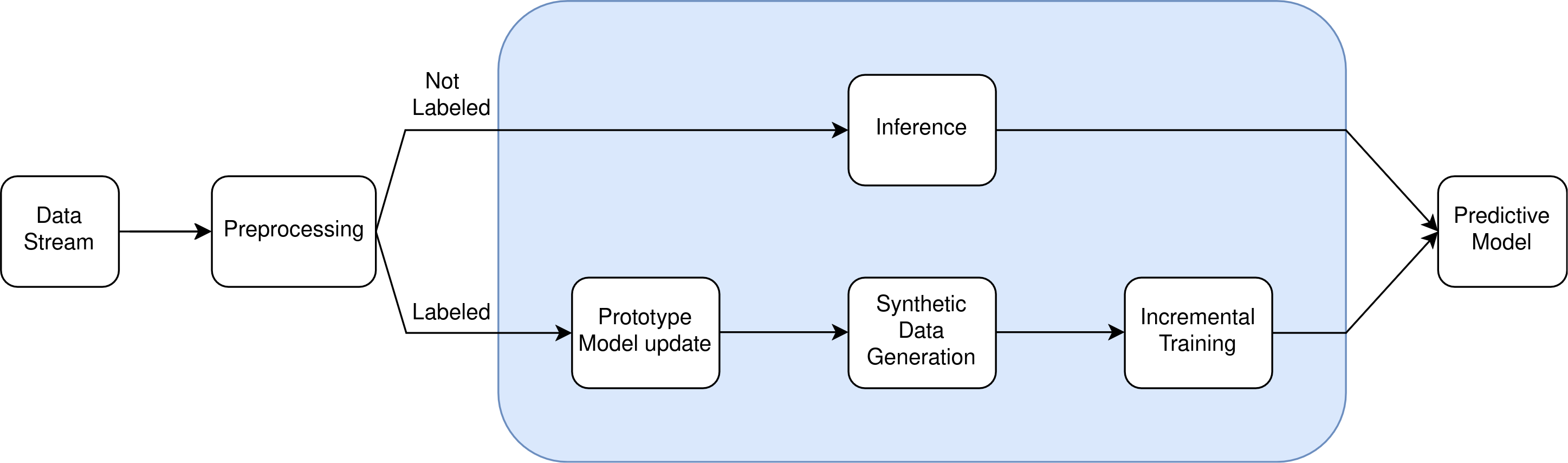}
  \caption{Diagram of the TRIL3 workflow.}

  \label{fig:simple}
\end{figure}

\begin{enumerate} 

\item \textbf{Initialization}: The DNDF and XuILVQ models are both set up. DNDF is initialized with random weights, whereas XuILVQ starts with an empty prototype set.

\item \textbf{Data Preprocessing}: To guarantee consistent input for both models, incoming data batches are standardized online.
    
\item \textbf{Prototype Update}: The XuILVQ prototype set is updated using labeled data. When new patterns are found, new prototypes are created, or existing prototypes are modified.
    
\item \textbf{Batch Generation}: To create a training batch, synthetic samples are generated from XuILVQ prototypes and mixed with new data. These synthetic samples are pseudo-randomly selected from the prototype set in such a way that all classes are balanced in the training batch. One adjustable factor that influences the equilibrium between stability and plasticity is the proportion of synthetic to real data.

\item \textbf{Classifier Update}: The classifier model is retrained using one epoch of stochastic gradient descent on the combined batch, enabling it to learn new patterns while rehearsing old knowledge through synthetic samples. To avoid overfitting, the classification model is only retrained when new prototypes are added to the prototype set (i.e., when new patterns are found in the input data).
        
\item \textbf{Prediction}: The classifier model processes unlabeled data to produce predictions. \end{enumerate}

Experiments in the original TRIL3 framework showed that 50\% synthetic data typically produced the best results across a variety of classification datasets. However, the XuILVQ algorithm's dependence on discrete class labels poses fundamental difficulties when applying this method to regression tasks, as we will address in~\Cref{sec:methodology}.

\section{Methodology}
\label{sec:methodology}

In continual learning scenarios, regression models must adjust to changing data distributions while retaining performance on previously learned patterns. The original TRIL3 framework described in the previous section was designed for classification and consists of two main components: a prototype-based generative model and a classifier model. Adapting TRIL3 to regression presents three fundamental challenges:

\begin{enumerate}
    \item \textbf{Absence of discrete labels for prototype generation}: XuILVQ requires 
   discrete class labels to create prototypes and determine similarity between samples. In regression, the continuous nature of target variables means no such labels exist, making it unclear how to define prototype neighborhoods.
   
   \item \textbf{The need to preserve prior knowledge under non-stationary conditions}: In continual learning scenarios, the model must maintain performance on previously learned patterns while adapting to new data distributions. This is particularly challenging in regression, where identical input regions may be assigned different target values over time due to concept drift, making the definition of prototype identity even more challenging. 
   \item \textbf{Represent an evolving target distribution}: Regression models must capture and adapt to potentially complex and time-varying distributions of continuous. The relationship between inputs and outputs may shift, requiring a representation that can model uncertainty.
\end{enumerate}

To address these challenges, we introduce the following modifications to TRIL3 
while preserving its core strategy:

\begin{enumerate}
    \item \textbf{Decision Tree Regressor (DTR) for virtual discretization}:  We introduce a DTR that partitions the continuous output space into discrete regions (leaf nodes), generating "virtual labels".
    \item \textbf{Preservation of past knowledge in prototypes}: the introduction of virtual labels provides sufficient structural information for the prototype model to successfully maintain prototypes that represent previously learned regions of the regression space.
    \item \textbf{Mixture Density Networks (MDN) as the regression model}: We replace DNDF with MDN to model the conditional probability distribution of continuous targets. 
    
\end{enumerate}

As mentioned, to enable prototype-based learning in a regression setting, we introduced a decision tree regressor (DTR) as an intermediate abstraction mechanism. The DTR is used to learn a discrete partition of the continuous target space. Each input sample is associated with a leaf node of the tree, which defines a virtual label corresponding to a region of similar input-output behaviour. These virtual labels do not represent semantic classes and are not defined externally, but are dynamically induced from continuous regression targets and evolve as the distribution of data changes over time. By discretising the continuous space in this way, DTR provides virtual labels that allow the XuILVQ module to create prototypes preserving the continuous nature of the task. 

Formally, let $\mathcal{D}_t = \{(\mathbf{x}_i, y_i)\}_{i=1}^{N_t}$ denote the stream of data observed up to time $t$, where $\mathbf{x}_i \in \mathbb{R}^d$ and $y_i \in \mathbb{R}$. The Decision Tree Regressor defines a mapping
\begin{equation}
f_{\mathrm{DTR}} : \mathbb{R}^d \rightarrow \mathcal{L}_t,
\end{equation}
where $\mathcal{L}_t$ is the set of leaf nodes at time $t$. Each sample $\mathbf{x}_i$ is assigned to a leaf
\begin{equation}
\ell_i = f_{\mathrm{DTR}}(\mathbf{x}_i),
\end{equation}
which generates a virtual label $v_i \in \{1,\dots,K_t\}$, with $K_t = |\mathcal{L}_t|$. 

 The uncertain nature of real-world data distributions is difficult for traditional point estimation regression models to capture, as they only predict a deterministic outcome for each input. To overcome these constraints, a strategy based on Mixture Density Networks (MDN) \footnote{Our implementation is based on the MDN architecture from \cite{repoMDN}, adapted to support single-step incremental updates for online learning scenarios.}~\cite{bishop1994mixture} was chosen. The main advantage that MDNs offer over traditional regression models is that instead of predicting a single value $y$ from an input $x$, they provide the parameters of a Gaussian Mixture Model. Specifically, they estimate the mean, $\mu$, and standard deviation, $\sigma$, of $N$ normal distributions. Additionally, each Gaussian component $k$ has an associated coefficient $\pi_k$ representing the probability that the output is generated by the $k$-th component. Thus, the conditional probability density of $y$ given an input $x$ is formulated as

\begin{equation}
    p(y|x)=\sum_{k=1}^N \pi_k(x)\mathcal{N}(y|\mu_k(x),\sigma_k(x)^2).
\end{equation}

Three main factors led us to the choice of using MDN: (1) adequate modeling capability to capture multimodal distributions; (2) architectural simplicity allowing for easy online learning adaptation; and (3) ease of integration with the TRIL3 prototype-based framework. Rather than achieving state-of-the-art regression performance, our main goal is to demonstrate effective continual learning with catastrophic forgetting mitigation, which makes MDN's balance of capability and simplicity ideal for our objectives. However, our framework is agnostic to the algorithm used to solve the regression task, so any other gradient-based model could be easily integrated instead.

The implemented model employs two Multi-Layer Perceptron neural networks to estimate the mixture parameters from the input data. The first network generates $N$ outputs corresponding to the mixture coefficients $\pi_k$, $k=1,\dots,N$. A log-softmax function is applied to these outputs to ensure numerical stability and guarantee that the coefficients sum to one. The other network computes the component parameters, namely the $N$ means $\mu_k$, $k = 1,\dots,N$ and the $N$ standard deviations $\sigma_k$, $k = 1,\dots,N$.

The model is trained by maximizing the data likelihood, which is equivalent to minimizing the negative log-likelihood (NLL) loss function. For a data pair $(x,y)$, the loss is calculated as:
\begin{equation}    
    \mathcal{L}(x,y) = -\log\left(\sum_{k=1}^N \pi_k(x)\mathcal{N}(y|\mu_k(x),\sigma_k(x)^2)\right).
\end{equation}

To obtain point predictions from the probabilistic model, we compute the expected value of the mixture distribution as the weighted average of component means:

\begin{equation}
    \mathbb{E}[y|x]=\sum_{k=1}^N \pi_k(x)\mu_k(x).
\end{equation}

Extending a continual learning framework based on prototypes, originally designed for classification, to regression poses challenges. In classification settings, prototypes are arranged around discrete class labels, which provide a stable supervision for prototype creation and adaptation. However, in regression, the absence of explicit class labels and the continuous nature of the target space mean that this mechanism cannot be used directly. In this context, it is unclear how to define the neighbourhoods of prototypes and how to decide whether new observations correspond to previously learned concepts. Furthermore, under non-stationary conditions, identical input regions may be assigned different target values over time, making the definition of prototype identity and similarity even harder.

\begin{figure}[H]
  \centering
    \includegraphics[width=1\textwidth]{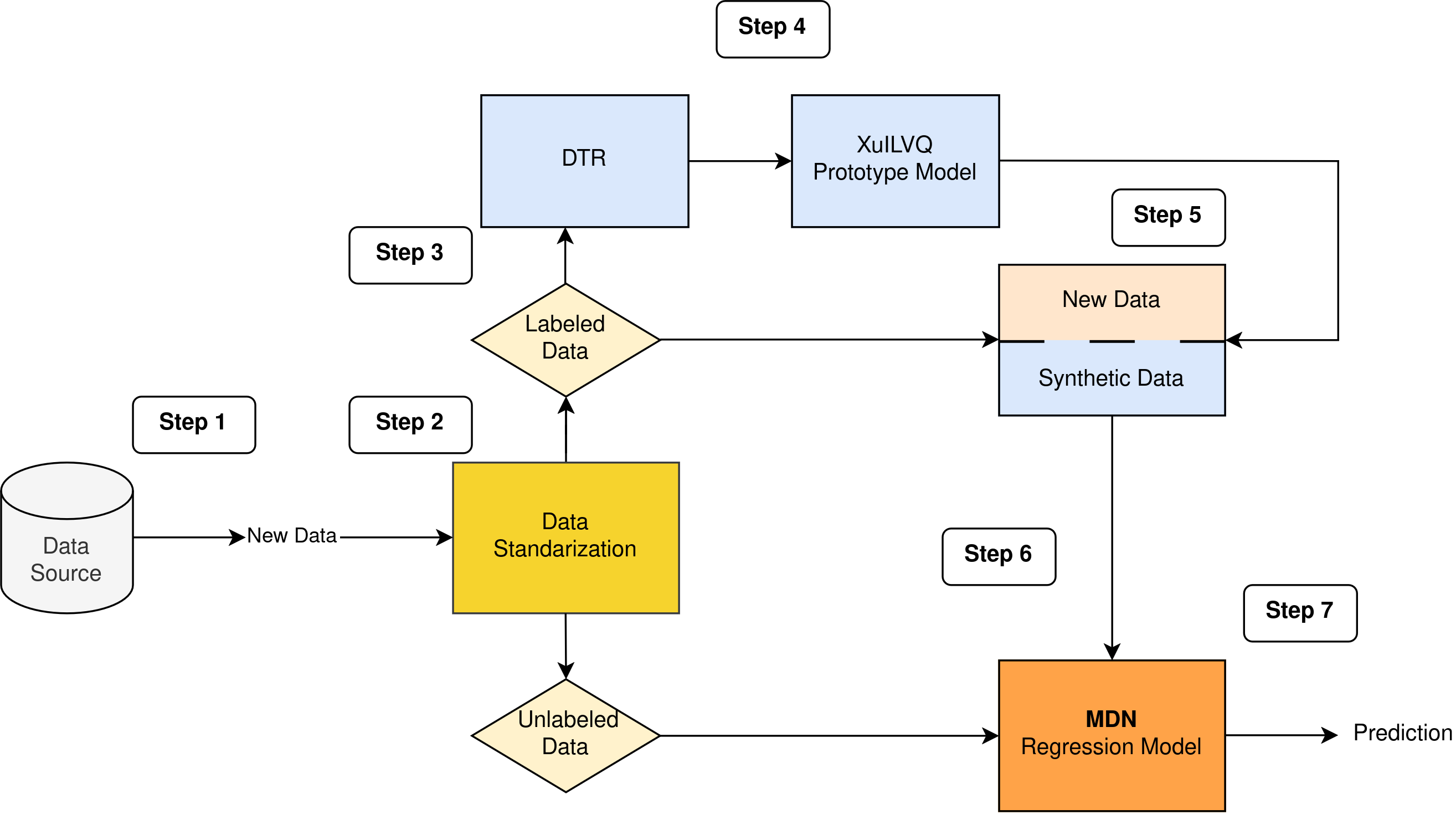}
  \caption{Diagram of the proposed continual regression framework.
}
  \label{fig:diagram}
\end{figure}

The workflow of our proposal, shown in~\autoref{fig:diagram} can be described as the following sequence of steps.

\begin{itemize}

\item [-] \textbf{Step 1.} In the first step, the incremental prototype model, the decision tree regressor, and the regression model are initialized. The prototype set maintained by XuILVQ is initialized as empty and grows dynamically as new data are observed. The framework then receives small batches of data in a continuous stream. These incoming batches may be labeled, i.e., associated with a target value, and so they will be used for training, or unlabeled, in which case they are used exclusively for prediction.
    
\item [-] \textbf{Step 2.} Each incoming batch is first transformed through online standardization to ensure numerical consistency across time. When labels are available, the batch contributes to each learning process: updating the regression model, the DTR model, and refining the prototype-based representation. In the absence of labels, the batch is not used for learning and is instead forwarded directly to the prediction stage.

\item [-] \textbf{Step 3.} The standardized batch is used to update the DTR. The role of the DTR is not to perform final prediction, but to induce a structured partition of the input space based on the continuous target variable. The DTR assigns each sample to a leaf node, which is later interpreted as a \emph{virtual label}. These virtual labels provide a discrete abstraction of the regression problem that enables prototype-based learning without relying on predefined classes.

\item [-] \textbf{Step 4.} Using the virtual labels inferred by the DTR, the XuILVQ prototype memory is updated. Incoming samples are compared against existing prototypes to determine whether they are adequately represented. New prototypes are created when the current prototype set fails to capture structural changes in the data distribution, while existing prototypes are incrementally adapted otherwise.  In addition, a control mechanism is provided to remove unused old prototypes, thus preventing the size of the entire prototype set from growing uncontrollably. This process allows the prototype memory to summarize past and present data in a compact and adaptive manner. It is important to highlight that continual labels, not virtual discrete ones from DTR, are stored after every update of the prototype set, thus a new training batch can be calculated during step 5 to effectively retrain the MDN regression model in a continual target space.

\item [-] \textbf{Step 5.} A rehearsal batch is then constructed by combining the newly observed real samples with synthetic samples generated from the prototype memory. To ensure balanced rehearsal, the continuous target space is discretized using quartiles derived from the prototype set, and synthetic samples are randomly selected to compensate for underrepresented quarters of this space.

\item [-] \textbf{Step 6.} The regression model is updated using the batch composed of real and synthetic data. This update reinforces knowledge acquired in earlier stages while allowing the model to adapt to new observations, thereby mitigating catastrophic forgetting in the presence of distributional shifts. To avoid overfitting the MDN regression model, steps 5 and 6 are only performed if the prototype set is updated during step 4 with new, unseen input patterns.

\item [-] \textbf{Step 7.} Finally, the updated regression model is used to generate predictions for incoming data. As new batches continue to arrive, the same sequence of steps is repeated.
\end{itemize}

\section{Validation and Results}
\label{sec:results}

In order to validate our proposal, we designed several complementary validation stages. On the one hand, our model is compared to an offline model to evaluate its accuracy, and then it is compared to another rehearsal technique to compare their performance against catastrophic forgetting (\Cref{subsec:phase1}). On the other hand, we compare our approach with the CLeaR approach~\cite{he2021clear}, allowing for a direct evaluation against a state-of-the-art regression framework under its own experimental assumptions (\Cref{subsec:clear_validation}). Another key aspect of approaches that use rehearsal techniques is their ability to manage memory efficiently in prolonged data flows. This aspect is relevant in prototype-based approaches, as the growth and evolution of the prototype set directly reflect how past knowledge is summarised. For this reason, a memory analysis is carried out to study the internal behaviour of prototype-based memory (\Cref{subsec:memoryanalysis}).

\begin{table}[htb]
\scriptsize
\centering
\begin{tabular}{lcccc}
\toprule
\textbf{Dataset} & \textbf{\# Samples} & \textbf{\# Features} & \textbf{Target Variable} \\
\midrule
California Housing Price & 20\,640 & 8 & House price \\
Fast Storage (Bitbrains) & 175\,950 & 8 & CPU usage (\%) \\
Wind Power Data & 47\,853 & 6 & Generated power \\
Solar Farm & 91\,813 & 6 & Generated power \\
Superconductors & 15\,457 & 81 & Critical temperature \\
Houses & 15\,004 & 8 & House price \\
House\_16H & 16\,552 & 16 & House price \\
Diamonds & 39\,106 & 9 & Diamond price \\
Bike Sharing & 12\,572 & 12 & Bike rentals \\
\bottomrule
\end{tabular}
\caption{Summary of the datasets used.}
\label{tab:datasets_summary}

\end{table}

First-stage experiments were conducted on multiple tabular regression datasets with heterogeneous characteristics in terms of dimensionality, scale, and target distribution. A little overview of all the datasets used can be seen in~\Cref{tab:datasets_summary}. Performance is evaluated using the Mean Squared Error (MSE) and the coefficient of determination ($R^2$) metrics, which are the most common ones in regression tasks.

\subsection{Phase 1: Evaluation Against Offline and Replay Baselines}
\label{subsec:phase1}

In the first phase, the objective is to evaluate the performance of the proposal against two baselines, one for accuracy and the other for capacity against catastrophic forgetting. To this end, the model will be compared in terms of predictive performance against an offline Random Forest Regressor(RFR)~\cite{Breiman2001}, given that the latter has unlimited access to samples, representing an upper-bound comparison. Due to have an incremental learning setting, the proposed model is not expected to match the accuracy of this offline counterpart. On the other hand, in continual learning systems, it is very important to measure their capacity against forgetting. Therefore, a replay method was also chosen as a baseline to compare the performance of our proposal in this regard.

The experiments are designed to evaluate the performance of the proposed method in a controlled continual learning scenario, which includes learning and forgetting phases. This way, a forgetting phase is introduced to validate robustness against distributional shifts during a controlled period. The training timeline is divided into three parts: an initial learning phase (0–40\%) where the model sees data from all distributions, a forgetting phase (40–70\%) where all samples with target values above the 70th percentile (calculated from the test set) are removed, and a final recovery phase (70–100\%) where the full data distribution is restored. By hiding these extreme values in the middle of training, it is possible to evaluate the model's performance against catastrophic forgetting. The hyperparameters used in the experiments can be found at~\Cref{tab:hyperparameters}

\begin{table}[htb]
\centering
\scriptsize
\renewcommand{\arraystretch}{1.15}
\newlength{\paramwidth}
\setlength{\paramwidth}{0.28\linewidth}
\begin{tabular}{p{\paramwidth}p{0.18\linewidth}}
\hline
\textbf{Hyperparameter} & \textbf{Value} \\
\hline
\end{tabular}
\vspace{10pt}
\begin{tabular}{p{\paramwidth}p{0.18\linewidth}}
\hline
\multicolumn{2}{c}{\textbf{Decision Tree Regressor (DTR)}} \\
\hline
Maximum depth & 4 \\
\hline
Minimum samples to split & 30 \\
\hline
Minimum samples per leaf & 10 \\
\hline
Minimum impurity decrease & 0.005 \\
\hline
\end{tabular}
\vspace{10pt}
\begin{tabular}{p{\paramwidth}p{0.18\linewidth}}
\noalign{\hrule height 0.5pt}
\hline
\multicolumn{2}{c}{\textbf{XuILVQ Prototype Model}} \\
\hline
Winner learning rate ($\alpha_{\text{winner}}$) & 0.9 \\
\hline
Runner-up learning rate ($\alpha_{\text{runner}}$) & 0.1 \\
\hline
Prototype age threshold (\texttt{AgeOld}) & 400 \\
\hline
Cleanup interval ($\gamma$) & 150 \\
\hline
Initial number of prototypes & 5 \\
\hline
\end{tabular}
\vspace{6pt}
\begin{tabular}{p{\paramwidth}p{0.18\linewidth}}
\noalign{\hrule height 0.5pt}
\hline
\multicolumn{2}{c}{\textbf{Mixture Density Network (MDN)}} \\
\hline
Number of mixture components & 5 \\
\hline
Number of hidden layers & 2 \\
\hline
Hidden layer dimension & 512 \\
\hline
Learning rate & 0.0005 \\
\hline
Batch size & 16 \\
\hline
\end{tabular}
\caption{Hyperparameters used for the DTR, XuILVQ, and MDN.}
\label{tab:hyperparameters}
\end{table}

\begin{figure}[htb]
  \centering
      \includegraphics[width=1\textwidth]{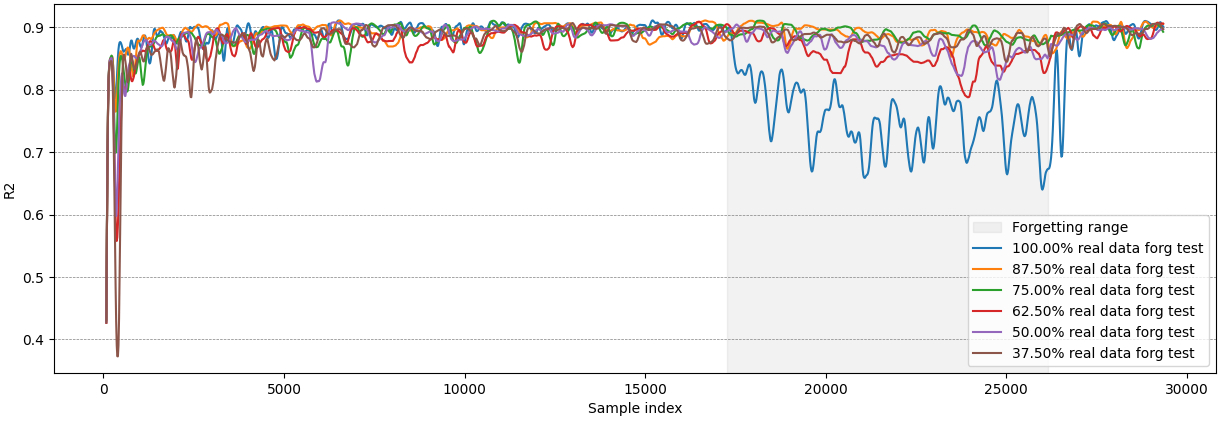}
  \caption{Evolution of R² in the diamonds dataset for different synthetic data ratios.}
  \label{fig:r2}
\end{figure}

In~\Cref{fig:r2}, the expected behaviour can be observed. In the configuration without synthetic data, the R2 decreases significantly, as it lacks the support of the prototype model. In contrast, in experiments that include synthetic data from prototypes, the R2 remains relatively stable during the forgetting phase.

\begin{figure}[htb]
  \centering
      \includegraphics[width=1\textwidth]{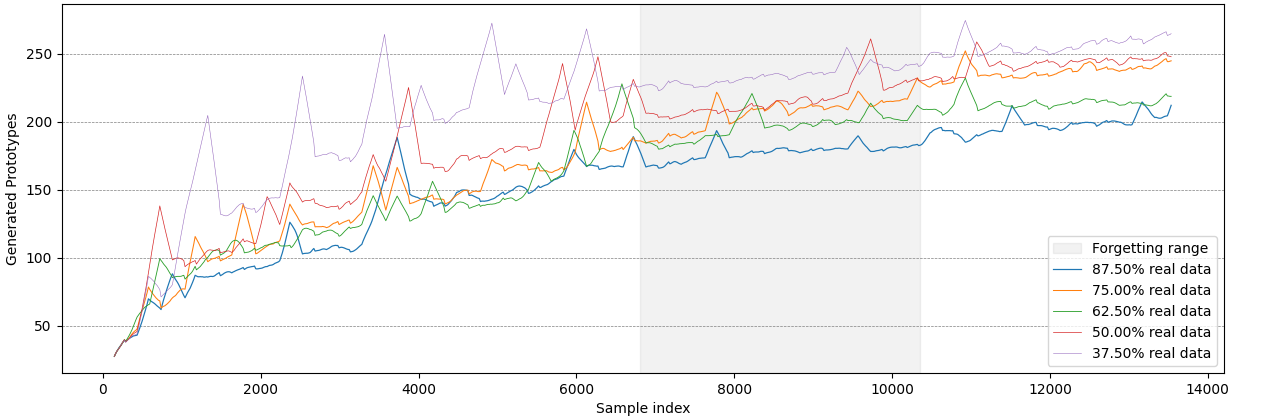}
  \caption{Evolution of the number of prototypes in the superconductors dataset  for different synthetic data ratios.}
  \label{fig:proto}
\end{figure}

As it is shown in ~\Cref{fig:proto}, the number of prototypes produced increases over time, initially sharply during the early stages of training, before stabilizing, suggesting a phase of exploration followed by the consolidation of prototypes that can be reused. The total number of prototypes produced correlates with the percentage of real data used in training; less use of real data results in more prototypes. Furthermore, during the forgetting phase, the prototype count remains stable, indicating effective memory maintenance and pruning. The conclusions are that the model adapts with controlled and stable growth over time.

For experiments comparing with the offline model, performance is presented using the final value obtained at the end of training. The metric is obtained after the forgetting phase and a recovery phase, where the previously lost class is recovered. For comparisons with replay, performance before forgetting is calculated as the average over a short window of the most recent training batches immediately preceding the start of forgetting. Specifically, the results are averaged over the last 20 batches to reduce the variability observed in instantaneous measurements. Performance during the forgetting phase is calculated as the average over the entire forgetting period.

\subsubsection{Comparison to Random Forest Regressor}

To establish a performance baseline, the proposed continual learning framework is validated against a Random Forest Regressor trained in a traditional offline batch setting. The RFR was selected as a benchmark due to its widely proven efficacy in regression tasks for tabular data. The dataset is partitioned into an 80\% training set and a 20\% test set. While the offline baseline is trained on the complete training corpus simultaneously, our proposal processes the data stream sequentially. After completion, both models are evaluated on the identical test split, using the MSE and the R2.

\subsubsection{Comparison to replay method}
\label{subsubsec:replay}

This experiment evaluates the robustness of the proposed method against catastrophic forgetting under a controlled non-stationary scenario. The experiments for this test consisted of simulating a continuous flow of data that, during a given period, does not include a specific distribution of the data, depriving the model of this knowledge. The procedure is executed with different ratios of synthetic data, both for 0\% synthetic data, without the use of prototypes, and for higher percentages, where the data generated by the prototype model would be added to the training samples. 

To perform the comparison, Avalanche Framework~\cite{JMLR:v24:23-0130} to was used to implement the replay strategy. The MDN architecture that was used to develop our model was also used to implement the replay method. Therefore, differences observed in performance can only be related to the continual learning strategies. The hyperparameters of the replay method can be seen at~\Cref{tab:replay_hyperparameters}.

\begin{table}[h]
\centering
\scriptsize
\renewcommand{\arraystretch}{1.2}
\begin{tabular}{lc}
\toprule
\textbf{Hyperparameter} & \textbf{Value} \\
\midrule

\multicolumn{2}{c}{\textbf{Replay Strategy}} \\
\midrule
Replay buffer size & 1000 \\

\midrule
\multicolumn{2}{c}{\textbf{MDN}} \\
\midrule
Number of hidden layers & 2 \\
Hidden layer dimension & 512 \\
Number of mixture components & 5 \\
Learning rate & 0.001 \\

\bottomrule
\end{tabular}
\caption{Hyperparameters used for the Avalanche Replay baseline.}
\label{tab:replay_hyperparameters}
\end{table}

To ensure that forgetting effects are properly isolated, evaluation is performed on a test set containing the suppressed samples. This prevents improvements in commonly observed data from masking performance degradation in infrequently observed target regions. For each configuration, three metrics are extracted: the average MSE before the forgetting phase, the average MSE during the forgetting phase, and a relative degradation index defined as
\[
\text{Degradation index} =
\frac{MSE_{\text{during}} - MSE_{\text{before}}}{MSE_{\text{before}}}.
\]

We used this metric to quantify performance degradation during the forgetting phase. It is defined as the normalized increase in mean squared error with respect to the pre-forgetting baseline.

\subsubsection{Phase 1 Results}

~\Cref{tab:offline_compact} reports the final performance obtained by the proposed method for different proportions of synthetic data, alongside the offline Random Forest baseline. The full table can be seen at ~\Cref{tab:resultados_rend_final}, the results shown in the table are an average of those obtained after 5 runs. Despite operating under much more restrictive conditions, our method achieves competitive performance compared to the offline Random Forest model. While our method processes data incrementally and strictly online, never returning to previous examples, the offline model represents an upper bound, as it has unlimited access to the entire dataset during training. The performance differences remain small on most benchmarks: datasets such as Diamonds (0.066 vs. 0.059 MSE), Wind Power (0.014 vs. 0.010 MSE) or Solar Farm (0.0061 vs. 0.0052 MSE) show that continual learning can approach and nearly match batch learning performance.

\begin{table}[htb]
\centering
\scriptsize
\begin{tabular}{lcccc}
\toprule
\textbf{Dataset} &
\multicolumn{2}{c}{\textbf{Our proposal (best config)}} &
\multicolumn{2}{c}{\textbf{Offline RFR}} \\
\cmidrule(lr){2-3} \cmidrule(lr){4-5}
 & \textbf{MSE} & $R^2$ & \textbf{MSE} & $R^2$ \\
\midrule
California Housing & 0.421 & 0.678 & 0.253 & 0.807 \\
Fast Storage       & 98.42& 0.695 & 43.19 & 0.866 \\
Houses             & 0.088 & 0.732 & 0.053 & 0.838 \\
Wind Power         & 0.014 & 0.553 & 0.010 & 0.669 \\
Wine Quality       & 0.529 & 0.319 & 0.345 & 0.555 \\
Solar Farm         & 0.0061 & 0.83 & 0.0052 & 0.854 \\
Superconductors    & 220.6 & 0.808 & 80.56 & 0.930 \\
House\_16H         & 0.469 & 0.395 & 0.321 & 0.588 \\
Bike Sharing       & 15430 & 0.514 & 10259.6 & 0.676 \\
Diamonds           & 0.066 & 0.936 & 0.059 & 0.943 \\
Wave Energy        & 6.71e6 & 0.552 & 2.67e5 & 0.982 \\
\bottomrule
\end{tabular}
\caption{Final predictive performance of our proposal compared to an offline Random Forest baseline.}
\label{tab:offline_compact}
\end{table}

~\Cref{tab:replay2} summarizes the results for the proposed method under varying proportions of synthetic data and compares them against a baseline based on a replay strategy. As in the previous experiment, the data in our proposal is an average of five runs. The configuration without synthetic data (0\%) behaves as a naive incremental learner and consistently has severe degradation, confirming the presence of catastrophic forgetting. In contrast, our proposal configurations incorporating synthetic replay substantially mitigate forgetting across all datasets. For the majority of benchmarks, the proportion of 62.5\% synthetic data achieves very good results. Furthermore, in nine out of eleven datasets, our approach performs better than experience replay, providing a substitute with advantages in both memory consumption and privacy policies. This benefit is demonstrated by results from particular datasets, such as Wave Energy (-0.3578 vs. 0.6277 MSE) or Houses (-0.0056 vs. 0.478 MSE). 

\begin{table}[htb]
\centering
\scriptsize
\begin{tabular}{lcccccc|c}
\hline
                   & \multicolumn{6}{c|}{\textbf{Our proposal (Synthetic Data\%)}} & \textbf{Replay} \\ \cline{2-8}
\textbf{Dataset}   & 62.5 & 50 & 37.5 & 25 & 12.5 & 0 & -- \\ \hline
California Housing & 0.1055 & 0.1418 & \textbf{0.0718} & 0.1097 & 0.2049 & 0.6240 & 0.3418 \\
Fast Storage       & -0.0052 & 0.0153 & -0.0007 & 0.0271 & \textbf{-0.1487} & 0.2533 & 0.0354 \\
Houses             & \textbf{-0.0056} & 0.0221 & -0.0009 & 0.0401 & 0.1485 & 0.3389 & 0.4778 \\
Wind Power         & 0.1833 & 0.1948 & 0.1676 & 0.1875 & 0.3441 & 0.6670 & \textbf{0.1410} \\
Wine Quality       & \textbf{0.1007} & 0.2599 & 0.1391 & 0.2159 & 0.2953 & 0.4290 & 0.8300 \\
Solar Farm         & \textbf{0.0545} & 0.1162 & 0.0800 & 0.1735 & 0.3651 & 2.5869 & 0.2342 \\
Superconductors    & \textbf{0.1870} & 0.3211 & 0.7081 & 0.4730 & 0.7427 & 4.5566 & 0.8466 \\
House\_16H         & 0.1490 & 0.1316 & \textbf{0.1046} & 0.2387 & 0.3446 & 0.5807 & 0.3397 \\
Bike Sharing       & \textbf{0.1156} & 0.3417 & 0.1789 & 0.4928 & 0.4803 & 0.9027 & 0.2056 \\
Diamonds           & -0.0237 & \textbf{-0.0914} & -0.0190 & -0.0596 & 0.0424 & 0.4402 & 0.0089 \\
Wave Energy        & \textbf{-0.3578} & -0.1117 & -0.0648 & -0.0431 & -0.0388 & -0.1177 & 0.6277 \\ \hline
\end{tabular}
\caption{Comparison table of degradation index between our proposal and the replay method. Lower values indicate better robustness.}
\label{tab:replay2}
\end{table}

\subsection{Phase 2: Evaluation Against CLeaR}
\label{subsec:clear_validation}

In addition to the validation with the replay-based strategy and offline regressor baselines, the proposed method was evaluated under the experimental protocol introduced in CLeaR~\cite{he2021clear}.

 In the second phase, the objective is to evaluate the performance of the proposed framework against another continual learning protocol, comparing it directly with a state-of-the-art solution. Identifying suitable baselines for this comparison was not straightforward, as there are relatively few methods explicitly designed for regression tasks. This experiment followed the evaluation methodology and data streaming setup of the original CLeaR algorithm, including the separation into warm-up, update, and evaluation phases, the use of a single-pass non-stationary data stream, and the definition of the forgetting ratio based on the variation of the prediction error before and after the update phase. This choice allows for a controlled comparison under identical conditions. 

The evaluation is conducted on the Europe Wind Farm dataset~\cite{gensler2021europewindfarm}, following the same preprocessing steps described in the original work, including removing extended periods of zero production, and min-max normalizing the input features and target values. Each wind farm was treated as a separate stream of data, leading to 10 separate continual learning scenarios. For each stream, the model was completed offline during an initial training period until convergence, then updated in real-time through one-pass processing of data. Performance was finally assessed on a held-out evaluation set that was never observed during training. Predictive performance on unseen data is measured at the end of all learning. Second, robustness to catastrophic forgetting is evaluated using the forgetting ratio defined in CLeaR, which measures the relative increase in prediction error on the warm-up data after the update phase. Better retention of previously learned information is indicated by lower values.

In the original CLeaR framework, three instances are evaluated. Instance A corresponds to a static model trained only during an initial warm-up phase and never updated thereafter, thus representing a lower bound on performance. Instance B extends this configuration by updating the model during data flow using standard fine-tuning, without any explicit mechanism to preserve previously acquired knowledge, making it prone to catastrophic forgetting. Instance C further improves on instance B by incorporating online elastic weight consolidation (Online-EWC) during updates, regularizing parameter changes. All three instances share the same buffer management system: a finite novelty buffer that stores samples with high prediction errors and triggers model updates when it is full, and an infinite familiarity buffer that stores well-predicted samples used for validation after each update.

In the~\Cref{tab:clear_forgetting_comparison}, labels B and C correspond to the two update mechanisms already mentioned: Instance$_B$ denotes standard online fine-tuning using newly arriving data, while Instance$_C$ incorporates Online-EWC. The suffix \textit{ae} indicates the inclusion of an autoencoder module for representation learning, which is applied prior to the regression model. Instance${_{B,\mathrm{ae}}}$ and Instance${_{C,\mathrm{ae}}}$ follow the same update strategies as their respective base variants, but operate on latent representations learned by the autoencoder rather than on the raw input space.

The results show that the proposed method consistently matches or improves the performance of the CLeaR experiments. In terms of predictive accuracy, as can be seen in~\Cref{tab:clear_prediction_comparison}, our solution achieves the lowest prediction error in seven out of ten wind farms when compared with all CLeaR instances. With respect to catastrophic forgetting, our strategy demonstrates a clear advantage. As reported in the~\Cref{tab:clear_forgetting_comparison}, we have a lower forgetting ratio in nine out of ten wind farms, and in several cases, it eliminates performance degradation during the update phase. 

These results show that the proposed method not only has a good performance under a standardized continual learning evaluation protocol, but also provides substantially improved robustness to catastrophic forgetting compared to buffer-based replay strategies such as CLeaR. 

\begin{table}[htb]
\centering
\scriptsize
\begin{tabular}{lcccc}
\toprule
\textbf{Wind Farm} &
\textbf{CLeaR A} &
\textbf{CLeaR B} &
\textbf{CLeaR C} &
\textbf{Our proposal} \\
\midrule
WF1  & 0.0242 & 0.0233 & 0.0182 & \textbf{0.0129} \\
WF2  & 0.0074 & 0.0095 & \textbf{0.0052} & 0.0357 \\
WF3  & 0.0237 & 0.0537 & 0.0121 & \textbf{0.0080} \\
WF4  & 0.0212 & 0.0164 & 0.0222 & \textbf{0.0156} \\
WF5  & 0.1386 & 0.0567 & 0.0556 & \textbf{0.0343} \\
WF6  & 0.0861 & 0.0802 & 0.0437 & \textbf{0.0043} \\
WF7  & 0.0097 & 0.0073 & \textbf{0.0059} & 0.0297 \\
WF8  & 0.0613 & 0.0497 & 0.0293 & \textbf{0.0184} \\
WF9  & 0.0254 & 0.0456 & 0.0182 & \textbf{0.0119} \\
WF10 & 0.0138 & 0.0301 & \textbf{0.0074} & 0.0162 \\
\bottomrule
\end{tabular}
\caption{Prediction error comparison on the Europe Wind Farm dataset.}
\label{tab:clear_prediction_comparison}
\end{table}

\begin{table}[htb]
\centering
\scriptsize
\begin{tabular}{lccccc}
\toprule
\textbf{Wind Farm} &
\textbf{CLeaR$_{B,ae}$} &
\textbf{CLeaR$_{C,ae}$} &
\textbf{CLeaR$_B$} &
\textbf{CLeaR$_C$} &
\textbf{Our proposal} \\
\midrule
WF1  & 1.474 & 0.727 & 1.955 & 0.994 & \textbf{0.000} \\
WF2  & 0.018 & 0.140 & 1.707 & 0.745 & \textbf{0.000} \\
WF3  & 1.205 & 1.489 & 6.170 & 0.421 & \textbf{0.021} \\
WF4  & 2.817 & 2.407 & 3.371 & 0.785 & \textbf{0.000} \\
WF5  & \textbf{0.000} & \textbf{0.000} & 7.617 & 2.296 & 0.019 \\
WF6  & 0.945 & 0.237 & 0.969 & 0.046 & \textbf{0.000} \\
WF7  & 0.574 & 1.552 & 1.768 & 1.253 & \textbf{0.000} \\
WF8  & 0.588 & 1.585 & 5.716 & 2.999 & \textbf{0.000} \\
WF9  & 1.463 & 0.848 & 3.470 & 1.463 & \textbf{0.034} \\
WF10 & 4.937 & 2.726 & 2.754 & 0.609 & \textbf{0.112} \\
\bottomrule
\end{tabular}
\caption{Forgetting ratio comparison on the Europe Wind Farm dataset.}
\label{tab:clear_forgetting_comparison}
\end{table}

\subsection{Memory Analysis}
\label{subsec:memoryanalysis}

One of the main objectives of the proposed framework is to mitigate catastrophic forgetting while maintaining a scalable memory footprint. Unlike buffer-based replay approaches that store raw historical samples, the proposed method, is based on a prototype-based representation in which the XuILVQ model gradually learns a compact set of prototypes that summarize prior knowledge.~\Cref{tab:datos_prototipos} reports the total number of training samples processed for each evaluated dataset, the final number of prototypes retained by the system, and the resulting prototype-to-data ratio (P/D), which provides a measure of memory efficiency. P/D ratios are typically below 2\% and well below 0.5\% for large-scale datasets like Fast Storage, Solar Farm, and Wind Power. The number of stored prototypes is much smaller than the total number of observed samples, indicating that the memory footprint of the system does not grow linearly with the volume of incoming data, but instead stabilizes as the prototype representation converges. Higher P/D ratios can be observed in datasets with greater variability or complexity, such as Wine Quality and House\_16H. Nevertheless, even in these cases, the absolute number of prototypes remains limited and lower than the number of samples that would be required by replay strategies based on raw data storage. For reference, the replay-based baseline evaluated in~\Cref{subsubsec:replay} uses a fixed memory buffer of 1000 stored samples. Overall, the results reported in ~\Cref{tab:datos_prototipos} demonstrate that the proposed prototype-based memory provides an effective balance between representational capacity and memory efficiency, remaining independent of the total number of observed samples, which makes it particularly useful for long-term continual learning scenarios that have memory constraints.

\begin{table}
\centering
\scriptsize
\begin{tabular}{lccc}
\toprule
\textbf{Dataset} & \textbf{Number of Samples} & \textbf{Number of Prototypes} & \textbf{P/D Ratio (\%)} \\
\midrule
California Housing & 15\,004 & 308 & 2.05 \\
Fast Storage       & 175\,950 & 377 & 0.22 \\
Houses             & 15\,004 & 206 & 1.37 \\
Wind Power         & 34\,730 & 117 & 0.37 \\
Wine Quality       & 3\,141  & 174 & 5.55 \\
Solar Farm         & 91\,813 & 210 & 0.23 \\
Superconductors    & 15\,457 & 217 & 1.40 \\
House\_16H         & 16\,552 & 809 & 4.89 \\
Bike Sharing       & 12\,572 & 158 & 1.26 \\
Diamonds           & 39\,106 & 119 & 0.30 \\
Wave Energy        & 26\,156 & 269 & 1.03 \\
\bottomrule
\end{tabular}
\caption{Memory efficiency report}
\label{tab:datos_prototipos}

\end{table}

\section{Conclusions}
\label{sec:conclusions}

In this paper, we introduce a new methodology designed to address the problem of catastrophic forgetting in online regression tasks. Our approach combines an intermediate Decision Tree Regressor (DTR), used to induce adaptive virtual labels from the continuous stream of data, with the XuILVQ prototype-based incremental generative model to generate synthetic data that preserves old knowledge. This is then integrated with a Mixed Density Network (MDN) architecture adapted to online learning to address regression tasks in tabular data. 

A key strength of the framework lies in its model-agnostic nature, this study utilized a MDN model due to its simplicity, but another regressor model could have been used, the framework serves as a modular backbone. Our results demonstrate that even with a straightforward regressor, TRIL3 effectively mitigates catastrophic forgetting. This confirms that the core innovation provides a robust foundation that can be enhanced by integrating more complex models in future iterations.

The framework was validated in terms of its performance in resisting catastrophic forgetting, its predictive performance, and its efficient use of memory. In comparison with the CLeaR framework, our solution demonstrated superior stability, with lower forgetting rates in most test scenarios, eliminating performance degradation during update phases. Furthermore, compared to offline benchmarks and generic experience replay, our method achieved competitive predictive performance, matching or exceeding the accuracy of offline Random Forest models on several datasets. In addition, the amount of stored data does not grow uncontrollably. Experiments have shown that, in the long term, the number of prototypes tends to stabilise at a specific value which is generally low compared to the total amount of data processed. In addition, the system for deleting insignificant stored data helps to control the size of the memory.

Having demonstrated that the TRIL3 system is viable for regression, future work could explore more elaborate regression model alternatives to improve the general predictive capacity and optimize performance metrics beyond the current MDN implementation. Also, while current results on different benchmarks are promising, it is important to verify the system's operation in a real-world scenario.

% Acknowledgments
\section*{Acknowledgment}
This work was supported by the grant PID2023-148716OB-C31 funded by MCIU/AEI/10.13039/501100011033 (DISCOVERY project),  by the Slovak Recovery and Resilience Plan under the project Smart Data Pipelines for the Cognitive Compute Continuum (SPICE), Agreement No. 09I02-03-V01-00012/2025/VA/PZ. Additionally, it also has been funded by the Galician Regional Government under project ED431B 2024/41 (GPC). Funding for open access charge: Universidade de Vigo/CISUG.

% Bibliography
\bibliographystyle{IEEEtran}
\bibliography{cas-refs}

% Appendix
\appendices
\clearpage
\section{Detailed Experimental Results}
\label{app:detailed_results}
This appendix details the experimental results.~\Cref{tab:resultados_rend_final} presents predictive performance metrics, including Mean Squared Error (MSE) and R², showing the comparison with an offline Random Forest Regressor.~\Cref{tab:results_forg} summarizes the forgetting ratios, including the comparison with the replay-based baseline. Finally,~\Cref{tab:appendix_clear_prediction_rot} showcases the prediction error ratios for various CLeaR instances. The results confirm consistent trends, highlighting that prototype-based generative replay effectively mitigates catastrophic forgetting in continual learning contexts.

\begin{table}[H]
\centering
\small
\resizebox{\textwidth}{!}{
\begin{tabular}{@{}lccccccc@{}}
\toprule
& \multicolumn{6}{c}{\textbf{Our proposal}} & \textbf{Offline} \textbf{RFR} \\
\cmidrule(lr){2-7} \cmidrule(l){8-8}
\textbf{Synthetic data (\%)} & \textbf{62.5} & \textbf{50} & \textbf{37.5} & \textbf{25} & \textbf{12.5} & \textbf{0} & -- \\
\midrule

\multicolumn{8}{l}{\textbf{California Housing}} \\
MSE  & 0.5114 $\pm$ 0.014 & 0.4861 $\pm$ 0.022 & 0.4858 $\pm$ 0.037 & 0.4850 $\pm$ 0.039 & 0.5032 $\pm$ 0.052 & 0.4214 $\pm$ 0.0042 & 0.2533 \\
$R^2$ & 0.6098 $\pm$ 0.011 & 0.6290 $\pm$ 0.017 & 0.6293 $\pm$ 0.028 & 0.6299 $\pm$ 0.030 & 0.6160 $\pm$ 0.040 & 0.6784 $\pm$ 0.0032 & 0.8067 \\
\midrule

\multicolumn{8}{l}{\textbf{Fast Storage}} \\
MSE  & 105.4 $\pm$ 5.5 & 101.3 $\pm$ 2.7 & 109.1 $\pm$ 6.9 & 98.2 $\pm$ 3.9 & 108.4 $\pm$ 6.1 & 99.6 $\pm$ 4.7 & 43.1945 \\
$R^2$ & 0.6704 $\pm$ 0.016 & 0.6841 $\pm$ 0.008 & 0.6579 $\pm$ 0.021 & 0.6947 $\pm$ 0.012 & 0.6606 $\pm$ 0.018 & 0.6876 $\pm$ 0.015 & 0.8656 \\
\midrule

\multicolumn{8}{l}{\textbf{Houses}} \\
MSE  & 0.1004 $\pm$ 0.005 & 0.1036 $\pm$ 0.006 & 0.1038 $\pm$ 0.005 & 0.1008 $\pm$ 0.004 & 0.0945 $\pm$ 0.004 & 0.0882 $\pm$ 0.003 & 0.0526 \\
$R^2$ & 0.6921 $\pm$ 0.018 & 0.6832 $\pm$ 0.021 & 0.6826 $\pm$ 0.019 & 0.6904 $\pm$ 0.017 & 0.7158 $\pm$ 0.016 & 0.7319 $\pm$ 0.015 & 0.8379 \\
\midrule

\multicolumn{8}{l}{\textbf{Wind Power}} \\
MSE  & 0.0171 $\pm$ 0.0008 & 0.0166 $\pm$ 0.0007 & 0.0175 $\pm$ 0.0009 & 0.0170 $\pm$ 0.0008 & 0.0169 $\pm$ 0.0007 & 0.0141 $\pm$ 0.0006 & 0.0104 \\
$R^2$ & 0.4583 $\pm$ 0.041 & 0.4715 $\pm$ 0.038 & 0.4462 $\pm$ 0.045 & 0.4569 $\pm$ 0.042 & 0.4597 $\pm$ 0.039 & 0.5532 $\pm$ 0.036 & 0.6692 \\
\midrule

\multicolumn{8}{l}{\textbf{Wine Quality}} \\
MSE  & 0.5669 $\pm$ 0.028 & 0.6014 $\pm$ 0.032 & 0.5926 $\pm$ 0.030 & 0.5483 $\pm$ 0.027 & 0.5621 $\pm$ 0.029 & 0.5294 $\pm$ 0.026 & 0.3448 \\
$R^2$ & 0.2694 $\pm$ 0.034 & 0.2236 $\pm$ 0.037 & 0.2351 $\pm$ 0.035 & 0.2928 $\pm$ 0.033 & 0.2756 $\pm$ 0.034 & 0.3185 $\pm$ 0.032 & 0.5548 \\
\midrule

\multicolumn{8}{l}{\textbf{Solar Farm}} \\
MSE  & 0.0071 $\pm$ 0.0004 & 0.0083 $\pm$ 0.0005 & 0.0074 $\pm$ 0.0004 & 0.0068 $\pm$ 0.0003 & 0.0066 $\pm$ 0.0003 & 0.0061 $\pm$ 0.0002 & 0.0052 \\
$R^2$ & 0.7961 $\pm$ 0.022 & 0.7654 $\pm$ 0.025 & 0.7893 $\pm$ 0.023 & 0.8071 $\pm$ 0.021 & 0.8137 $\pm$ 0.020 & 0.8298 $\pm$ 0.019 & 0.8538 \\
\midrule

\multicolumn{8}{l}{\textbf{Superconductors}} \\
MSE  & 236.9 $\pm$ 14 & 262.4 $\pm$ 17 & 230.8 $\pm$ 13 & 232.1 $\pm$ 14 & 220.6 $\pm$ 12 & 234.9 $\pm$ 15 & 80.5583 \\
$R^2$ & 0.7951 $\pm$ 0.012 & 0.7703 $\pm$ 0.014 & 0.8002 $\pm$ 0.011 & 0.7991 $\pm$ 0.012 & 0.8084 $\pm$ 0.010 & 0.7960 $\pm$ 0.013 & 0.9300 \\
\midrule

\multicolumn{8}{l}{\textbf{House\_16H}} \\
MSE  & 0.5061 $\pm$ 0.023 & 0.4984 $\pm$ 0.021 & 0.5017 $\pm$ 0.022 & 0.5002 $\pm$ 0.023 & 0.5136 $\pm$ 0.025 & 0.4693 $\pm$ 0.020 & 0.3210 \\
$R^2$ & 0.3521 $\pm$ 0.029 & 0.3617 $\pm$ 0.028 & 0.3564 $\pm$ 0.029 & 0.3581 $\pm$ 0.028 & 0.3439 $\pm$ 0.030 & 0.3946 $\pm$ 0.027 & 0.5880 \\
\midrule

\multicolumn{8}{l}{\textbf{Bike Sharing}} \\
MSE  & 19870 $\pm$ 2100 & 19340 $\pm$ 2000 & 16890 $\pm$ 1800 & 18920 $\pm$ 1900 & 20110 $\pm$ 2200 & 15430 $\pm$ 1600 & 10259.58 \\
$R^2$ & 0.3718 $\pm$ 0.061 & 0.3854 $\pm$ 0.059 & 0.4682 $\pm$ 0.054 & 0.4021 $\pm$ 0.057 & 0.3657 $\pm$ 0.062 & 0.5136 $\pm$ 0.050 & 0.6760 \\
\midrule

\multicolumn{8}{l}{\textbf{Diamonds}} \\
MSE  & 0.0675 $\pm$ 0.0019 & 0.0684 $\pm$ 0.0021 & 0.0679 $\pm$ 0.0020 & 0.0713 $\pm$ 0.0024 & 0.0658 $\pm$ 0.0018 & 0.0671 $\pm$ 0.0019 & 0.0589 \\
$R^2$ & 0.9341 $\pm$ 0.006 & 0.9330 $\pm$ 0.006 & 0.9336 $\pm$ 0.006 & 0.9309 $\pm$ 0.007 & 0.9359 $\pm$ 0.005 & 0.9344 $\pm$ 0.006 & 0.9428 \\
\midrule

\multicolumn{8}{l}{\textbf{Wave Energy}} \\
MSE  & 7.91e6 $\pm$ 1.5e6 & 6.90e6 $\pm$ 9.9e5 & 6.88e6 $\pm$ 9.1e5 & 7.48e6 $\pm$ 8.4e5 & 6.95e6 $\pm$ 5.5e5 & 6.71e6 $\pm$ 9.1e5 & 2.67e5 \\
$R^2$ & 0.471 $\pm$ 0.098 & 0.539 $\pm$ 0.066 & 0.540 $\pm$ 0.061 & 0.500 $\pm$ 0.056 & 0.535 $\pm$ 0.037 & 0.552 $\pm$ 0.061 & 0.9821 \\
\bottomrule
\end{tabular}
}
\caption{Comparison of performance between our proposal and an offline Random Forest Regressor}
\label{tab:resultados_rend_final}
\end{table}

\begin{table}[t]
\centering
\small
\setlength{\tabcolsep}{4pt}
\resizebox{\textwidth}{!}{
\begin{tabular}{@{}lccccccc@{}}
\toprule
& \multicolumn{6}{c}{\textbf{Our proposal}} & \textbf{Replay} \\
\cmidrule(lr){2-7} \cmidrule(l){8-8}
\textbf{Synthetic data (\%)} & \textbf{62.5} & \textbf{50} & \textbf{37.5} & \textbf{25} & \textbf{12.5} & \textbf{0} & -- \\
\midrule

\multicolumn{8}{l}{\textbf{California Housing}} \\
Before & 0.908 $\pm$ 0.057 & 0.8315 $\pm$ 0.060 & 0.8215 $\pm$ 0.053 & 0.8682 $\pm$ 0.12 & 0.8206 $\pm$ 0.049 & \textbf{0.722 $\pm$ 0.027} & 0.7321 \\
During & 1.004 $\pm$ 0.091 & 0.9494 $\pm$ 0.044 & \textbf{0.8805 $\pm$ 0.0094} & 0.9635 $\pm$ 0.094 & 0.9887 $\pm$ 0.046 & 1.173 $\pm$ 0.020 & 0.9823 \\
Degradation & 0.1055 & 0.1418 & \textbf{0.0718} & 0.1097 & 0.2049 & 0.6240 & 0.3418 \\
\midrule

\multicolumn{8}{l}{\textbf{Fast Storage}} \\
Before & \textbf{238.2 $\pm$ 24} & 238.7 $\pm$ 5.7 & 244.8 $\pm$ 30 & 233.5 $\pm$ 10 & 286.7 $\pm$ $10^2$ & 249.3 $\pm$ 22 & 298.0 \\
During & \textbf{237.0 $\pm$ 8.8} & 242.4 $\pm$ 8.8 & 244.6 $\pm$ 12 & 239.8 $\pm$ 8.3 & 244.0 $\pm$ 11 & 312.5 $\pm$ $10^2$ & 308.6 \\
Degradation & {-0.0052} & 0.0153 & -0.0007 & 0.0271 & \textbf{-0.1487} & 0.2533 & 0.0354 \\
\midrule

\multicolumn{8}{l}{\textbf{Houses}} \\
Before & 0.1085 $\pm$ 0.0049 & 0.1083 $\pm$ 0.0047 & 0.1078 $\pm$ 0.0044 & 0.1096 $\pm$ 0.0061 & 0.1127 $\pm$ 0.0060 & 0.1159 $\pm$ 0.0048 & \textbf{0.0900} \\
During & \textbf{0.1072 $\pm$ 0.0058} & 0.1107 $\pm$ 0.0057 & 0.1084 $\pm$ 0.0034 & 0.1140 $\pm$ 0.0053 & 0.1294 $\pm$ 0.0062 & 0.1552 $\pm$ 0.0075 & 0.1330 \\
Degradation & \textbf{-0.0056} & 0.0221 & -0.0009 & 0.0401 & 0.1485 & 0.3389 & 0.4778 \\
\midrule

\multicolumn{8}{l}{\textbf{Wind Power}} \\
Before & 0.0453 $\pm$ 0.0044 & 0.0426 $\pm$ 0.0039 & 0.0418 $\pm$ 0.0037 & 0.0432 $\pm$ 0.0038 & 0.0401 $\pm$ 0.0029 & 0.0376 $\pm$ 0.0021 & \textbf{0.0359} \\
During & 0.0536 $\pm$ 0.0045 & 0.0509 $\pm$ 0.0024 & 0.0488 $\pm$ 0.0026 & 0.0513 $\pm$ 0.0032 & 0.0539 $\pm$ 0.0031 & 0.0627 $\pm$ 0.0034 & \textbf{0.0403} \\
Degradation & 0.1833 & 0.1948 & 0.1676 & 0.1875 & 0.3441 & 0.6670 & \textbf{0.1410} \\
\midrule

\multicolumn{8}{l}{\textbf{Wine Quality}} \\
Before & 0.5140 $\pm$ 0.031 & 0.5216 $\pm$ 0.036 & 0.5056 $\pm$ 0.032 & 0.5123 $\pm$ 0.034 & 0.5034 $\pm$ 0.028 & 0.4981 $\pm$ 0.030 & \textbf{0.3956} \\
During & 0.5657 $\pm$ 0.038 & 0.6571 $\pm$ 0.045 & \textbf{0.5759 $\pm$ 0.025} & 0.6228 $\pm$ 0.041 & 0.6519 $\pm$ 0.039 & 0.7116 $\pm$ 0.046 & 0.7239 \\
Degradation & \textbf{0.1007} & 0.2599 & 0.1391 & 0.2159 & 0.2953 & 0.4290 & 0.8300 \\
\midrule

\multicolumn{8}{l}{\textbf{Solar Farm}} \\
Before & 0.0202 $\pm$ 0.0014 & 0.0198 $\pm$ 0.0011 & 0.0200 $\pm$ 0.0012 & 0.0196 $\pm$ 0.0010 & 0.0189 $\pm$ 0.0011 & 0.0184 $\pm$ 0.0010 & \textbf{0.0176} \\
During & 0.0213 $\pm$ 0.0011 & 0.0221 $\pm$ 0.0010 & 0.0216 $\pm$ 0.0010 & 0.0230 $\pm$ 0.0011 & \textbf{0.0258 $\pm$ 0.0014} & 0.0660 $\pm$ 0.0035 & 0.0217 \\
Degradation & \textbf{0.0545} & 0.1162 & 0.0800 & 0.1735 & {0.3651} & 2.5869 & 0.2342 \\
\midrule

\multicolumn{8}{l}{\textbf{Superconductors}} \\
Before & 451.3 $\pm$ 29 & 417.1 $\pm$ 28 & 404.7 $\pm$ 33 & 421.9 $\pm$ 31 & 438.5 $\pm$ 35 & 406.6 $\pm$ 27 & \textbf{372.7} \\
During & \textbf{480.3 $\pm$ 35} & 551.0 $\pm$ 49 & 770.8 $\pm$ 68 & 621.6 $\pm$ 54 & 764.3 $\pm$ 72 & 2259.4 $\pm$ 110 & 688.3 \\
Degradation & \textbf{0.1870} & 0.3211 & 0.7081 & 0.4730 & 0.7427 & 4.5566 & 0.8466 \\
\midrule

\multicolumn{8}{l}{\textbf{House\_16H}} \\
Before & 0.2297 $\pm$ 0.020 & 0.2284 $\pm$ 0.019 & 0.2266 $\pm$ 0.020 & 0.2231 $\pm$ 0.018 & 0.2184 $\pm$ 0.017 & \textbf{0.2028 $\pm$ 0.015} & 0.3080 \\
During & \textbf{0.2503 $\pm$ 0.016} & 0.2585 $\pm$ 0.019 & 0.2639 $\pm$ 0.020 & 0.2764 $\pm$ 0.021 & 0.2937 $\pm$ 0.023 & 0.3206 $\pm$ 0.025 & 0.4127 \\
Degradation & \textbf{0.1046} & 0.1316 & 0.1490 & 0.2387 & 0.3446 & 0.5807 & 0.3397 \\
\midrule

\multicolumn{8}{l}{\textbf{Bike Sharing}} \\
Before & 36130 $\pm$ 6400 & 37790 $\pm$ 7200 & 36210 $\pm$ 6800 & 39360 $\pm$ 7100 & 42110 $\pm$ 7500 & \textbf{32930 $\pm$ 5900} & 35507.7 \\
During & \textbf{40310 $\pm$ 6500} & 50720 $\pm$ 8600 & 42690 $\pm$ 5100 & 58740 $\pm$ 9200 & 62310 $\pm$ 9800 & 62650 $\pm$ 10200 & 42809.8 \\
Degradation & \textbf{0.1156} & 0.3417 & 0.1789 & 0.4928 & 0.4803 & 0.9027 & 0.2056 \\
\midrule

\multicolumn{8}{l}{\textbf{Diamonds}} \\
Before & 0.0794 $\pm$ 0.006 & 0.08268 $\pm$ 0.011 & 0.08214 $\pm$ 0.011 & 0.08085 $\pm$ 0.0094 & 0.07374 $\pm$ 0.0079 & \textbf{0.06931 $\pm$ 0.0012} & 0.0873 \\
During & \textbf{0.07512 $\pm$ 0.0045} & 0.07751 $\pm$ 0.0073 & 0.08058 $\pm$ 0.0071 & 0.07603 $\pm$ 0.0050 & 0.07687 $\pm$ 0.0044 & 0.09983 $\pm$ 0.0025 & 0.0881 \\
Degradation & \textbf{-0.0914} & -0.0237 & -0.0190 & -0.0596 & 0.0424 & 0.4402 & 0.0089 \\
\midrule

\multicolumn{8}{l}{\textbf{Wave Energy}} \\
Before & 2.54e6 $\pm$ 3.7e5 & 2.54e6 $\pm$ 5.1e5 & 3.67e6 $\pm$ 2.9e6 & 2.40e6 $\pm$ 1.6e5 & 2.37e6 $\pm$ 9.6e4 & 3.00e6 $\pm$ 6.4e5 & \textbf{1.41e6} \\
During & 2.38e6 $\pm$ 3.7e5 & 2.26e6 $\pm$ 3.4e5 & 2.35e6 $\pm$ 4.9e5 & 2.30e6 $\pm$ 2.9e5 & \textbf{1.99e6} & 2.65e6 $\pm$ 3.3e5 & 2.29e6 \\
Degradation & \textbf{-0.3578} & -0.1117 & -0.0648 & -0.0431 & -0.0388 & -0.1177 & 0.6277 \\
\bottomrule
\end{tabular}
}
\caption{Comparison of forgetting between our proposal and the replay strategy}
\label{tab:results_forg}
\end{table}

\begin{table}[h]
\centering
\small
\begin{tabular}{lcccccccccc}
\toprule
\textbf{Method (Ratio)} &
\textbf{WF1} & \textbf{WF2} & \textbf{WF3} & \textbf{WF4} & \textbf{WF5} &
\textbf{WF6} & \textbf{WF7} & \textbf{WF8} & \textbf{WF9} & \textbf{WF10} \\
\midrule

CLeaR-A      & 0.0321 & 0.0114 & 0.0196 & 0.0287 & 0.0612 & 0.0584 & 0.0143 & 0.0415 & 0.0269 & 0.0182 \\
CLeaR-B      & 0.0254 & 0.0079 & 0.0151 & 0.0219 & 0.0531 & 0.0497 & 0.0101 & 0.0342 & 0.0213 & 0.0126 \\
CLeaR-C      & 0.0182 & \textbf{0.0052} & 0.0121 & 0.0164 & 0.0457 & 0.0437 & \textbf{0.0059} & 0.0293 & 0.0181 & \textbf{0.0074} \\
\midrule
Our proposal (100\%)  & 0.0146 & 0.0412 & \textbf{0.0080} & 0.0171 & 0.0389 & \textbf{0.0043} & 0.0312 & 0.0195 & 0.0138 & 0.0179 \\
Our proposal (87.5\%) & 0.0142 & 0.0396 & 0.0098 & 0.0168 & 0.0375 & 0.0061 & 0.0304 & 0.0201 & 0.0145 & 0.0184 \\
Our proposal (75\%)   & 0.0139 & 0.0357 & 0.0094 & \textbf{0.0156} & 0.0361 & 0.0058 & 0.0300 & \textbf{0.0184} & 0.0142 & 0.0168 \\
Our proposal (62.5\%) & \textbf{0.0129} & 0.0381 & 0.0100 & 0.0163 & 0.0354 & 0.0051 & 0.0298 & 0.0189 & \textbf{0.0119} & 0.0172 \\
Our proposal (50\%)   & 0.0134 & 0.0420 & 0.0160 & 0.0175 & \textbf{0.0343} & 0.0069 & \textbf{0.0297} & 0.0198 & 0.0131 & \textbf{0.0162} \\
Our proposal (37.5\%) & 0.0141 & 0.0408 & 0.0100 & 0.0179 & 0.0348 & 0.0056 & 0.0309 & 0.0204 & 0.0146 & 0.0175 \\
\bottomrule
\end{tabular}
\caption{Prediction error under the CLeaR experimental protocol. }

\label{tab:appendix_clear_prediction_rot}
\end{table}

\end{document}